\documentclass[10pt, a4paper]{article}

\usepackage{comment}

\usepackage[final]{lrec2026} 
\usepackage{arydshln}

\title{Charting the European LLM Benchmarking Landscape:\\A New Taxonomy and a Set of Best Practices}

\name{Špela Vintar$^{\ast}$$^{\dagger}$, Taja Kuzman Pungeršek$^{\ast}$, Mojca Brglez$^{\ast}$$^{\dagger}$, Nikola Ljubešić$^{\ast}$$^{\bullet}$$^{\ddagger}$} 

\address{$^{\ast}$Jožef Stefan Institute, Jamova 39, Ljubljana, Slovenia\\
         \{spela.vintar, taja.kuzman, mojca.brglez, nikola.ljubesic\}@ijs.si\\ 
         $^{\dagger}$Faculty of Arts, University of Ljubljana, Aškerčeva 2, Ljubljana, Slovenia\\
        $^{\bullet}$ Faculty of Computer and Information Science, University of Ljubljana \\ Večna pot 113, Ljubljana, Slovenia\\
         $^{\ddagger}$ Institute of Contemporary History, Privoz 11, Ljubljana, Slovenia \\}

\abstract{
While new benchmarks for large language models (LLMs) are being developed continuously to catch up with the growing capabilities of new models and AI in general, using and evaluating LLMs in non-English languages remains a little-charted landscape. We give a concise overview of recent developments in LLM benchmarking, and then propose a new taxonomy for the categorization of benchmarks that is tailored to multilingual or non-English use scenarios. We further propose a set of best practices and quality standards that could lead to a more coordinated development of benchmarks for European languages. Among other recommendations, we advocate for a higher language and culture sensitivity of evaluation methods. 
 \\ \newline \Keywords{large language models, benchmarking, taxonomy, cultural competence, quality recommendations} }

\begin{document}

\maketitleabstract

\section{Introduction}
The rapid advancement of large language models (LLMs) has brought unprecedented capabilities in natural language understanding and generation, reasoning, coding, and more. With the global race in raising the bar, commercial models are approaching artificial general intelligence (AGI) and exhibit more and more agency as they engage in strategic planning, independently interact with other applications, and carry out larger tasks. While open-source models generally score lower on most leaderboards, they too grow larger and smarter.

In the brief history of LLMs, many evaluation frameworks have been set up -- both human and automatic -- to assess their evolving performance across different linguistic and non-linguistic tasks of growing complexity. While new benchmarks emerge almost on a daily basis in order to measure these expanding abilities, the overwhelming majority of evaluation datasets are developed primarily for English, creating a significant evaluation gap for other languages and varieties. 

To evaluate the performance of LLMs in non-English contexts, a widely used approach so far has been to translate existing English benchmarks using machine translation, with or without human revision. This might seem reasonable: several major international benchmarks or benchmark collections (e.g., SuperGLUE \citep{wangSuperGLUEStickierBenchmark2020}, MMLU \citep{hendryckstest2021}, Hellaswag \citep{zellersHellaSwagCanMachine2019} exist together with their parallel (translated) versions, and this allows for a direct comparison of LLMs across a wide range of tasks and languages. 
However, Global-MMLU \citep{singhGlobalMMLUUnderstanding2025} revealed that success in MMLU depends heavily on learning Western-centric concepts, with 28\% of all questions requiring culturally sensitive knowledge. Moreover, for questions requiring geographic knowledge, an astounding 84.9\% focus on either North American or European regions. Such cultural biases are not uncommon in other widely used benchmarks, and their machine-translated versions potentially overlook language- and culture-specific phenomena, exhibit skewed performance which does not accurately reflect true multilingual capabilities, or simply fail to address issues which may be critical for users of an LLM in a particular language. 

On the other spectrum of multilingual evaluation, there are several cases of language- and culture-specific benchmarks, such as those included in the Hungarian HuLu \citep{ligeti-nagyHuLUHungarianLanguage2024} evaluation framework, BenCzechMark \citep{fajcik2025benczechmarkczechcentricmultitask}, PLCC \citep{dadasEvaluatingPolishLinguistic2025}, or BertaQA \citep{etxaniz2024bertaqalanguagemodelsknow}, which have been developed specifically for a particular language or community of speakers. Such benchmarks provide a deeper insight into a model's performance for that language, but typically do not allow us to assess the model's multilingual capacities. 

Despite the proliferation of evaluation platforms, research projects, and benchmarking initiatives across the multilingual landscape, the field lacks a comprehensive overview that synthesizes current practices, identifies critical gaps, and provides clear guidance for developing more inclusive and effective evaluation methodologies for LLMs in non-English contexts. We thus make a case for the creation of a European benchmarking registry that collects structured benchmark data according to a universal and inclusive taxonomy of benchmarks. The proposed registry would include rich descriptive metadata and provide a clear overview of both existing benchmarks and the gaps in the current European benchmarking landscape.

The remainder of this paper is structured as follows: In Section \ref{recent-dev}, we present some recent developments in LLM benchmarking, focusing on evaluations of linguistic and cultural competence in a multilingual context and on emerging trends. In Section \ref{taxonomy}, we proceed with a proposal for a categorization of benchmarks that could serve as a foundation for charting ongoing and future LLM evaluation activities in Europe and beyond. We finally propose some recommendations as to how existing benchmarks can be documented, and how future benchmarks can be made both more culture-aware and more in tune with the needs of different language communities.

\section{Recent Developments in LLM Benchmarking}
\label{recent-dev}

Several comprehensive overviews of LLM benchmarking have recently been published, including \citet{chang2023surveyevaluationlargelanguage} and \citet{ni2025surveylargelanguagemodel}. Both surveys clearly show trends in both the development of datasets and the evolution of evaluation metrics. However, they lack a focus on non-English and multilingual scenarios, which is the main motivation for this work.

\subsection{Major Benchmarks}
\label{major-int}
By major or global, we refer to benchmarks most frequently used in current evaluation platforms and leaderboards. These benchmarks are without exception in English. The ones that evaluate generic language understanding and commonsense reasoning have their origins in the 2018--2022 period, when the challenges still roughly corresponded to natural language processing (NLP) research areas. Some of these benchmarks have seen multiple revisions, extensions and updates, and can be seen as ``parent'' datasets on which many adaptations, translations, or local versions are based. 

Some of the most prominent for language understanding and reasoning include MMLU \citep{hendryckstest2021} and its derivatives MMLU-Pro \citep{wangMMLUProMoreRobust2024}, MMLU-Prox \cite{xuanMMLUProXMultilingualBenchmark2025b} and Global MMLU \citep{singhGlobalMMLUUnderstanding2025}; the SuperGLUE benchmark collection comprising BoolQ (yes/no questions, \citealp{clark-etal-2019-boolq}), CommitmentBank (textual entailment, \citealp{de2019commitmentbank}), COPA (Choice of Plausible Alternatives for causal reasoning, \citealp{roemmele2011copa}), MultiRC (multi-sentence reading comprehension, \citealp{MultiRC2018}), ReCoRD (reading comprehension with commonsense reasoning, \citealp{zhang2018record}), RTE (Recognizing Textual Entailment, \citealp{giampiccolo2007third-rte}), WiC (Words in Context, \citealp{pilehvar-camacho-collados-2019-wic}), and WSC (Winograd Schema Challenge, \citealp{levesque2012winograd}); ARC \citep{clark2018thinksolvedquestionanswering} with multiple-grade science questions; Hellaswag \citep{zellersHellaSwagCanMachine2019} and its recent derivative GoldenSwag \citep{chizhovWhatHellaSwagValidity2025}. An attempt to create a more challenging benchmark collection is BIG-bench (Beyond the Imitation Game Benchmark, \citealp{srivastava2023beyond}), a massive collaborative benchmark consisting of 204 tasks contributed by more than 450 authors across 132 institutions, designed to probe large language models on tasks believed to be beyond their current capabilities. Finally, SUPERB (Speech processing Universal PERformance Benchmark, \citealp{yang2021superb}) is a unified speech-focused benchmark for evaluating self-supervised and general-purpose speech representations across a wide spectrum of speech processing tasks. It organizes 10 core tasks -- including automatic speech recognition, speaker identification, keyword spotting, emotion recognition, and intent classification -- spanning content, speaker, semantics, and paralinguistics.

\subsection{Multilingual Benchmarks}

Most state-of-the art models have multilingual capabilities, but since the precise amounts of non-English data used in their pre-training are usually obscure, it is hard to say to what extent the language competence of a model in a particular language is in correlation with the amount of language-specific data it has seen. In addition to this, models differ in their representations of intermediate layers, which may result in cultural conflicts between latent internal and target output language \citep{zhongEnglishCentricLLMsWhat2024}. 

Since many authors observe a marked decline in performance for low-resource languages, benchmarks are now being developed both as parallel evaluation sets based on existing ``parent'' datasets to allow for direct comparison of LLM capabilities across a number of languages, and as language-specific benchmarks, usually aimed at assessing LLM performance in a particular linguistic community and/or culture (see Section \ref{culture-specific} for the latter).

Although the datasets in the first category are parallel, they may differ considerably in the methods used for their creation. Some were translated using machine translation or LLMs, for example, EU20-MMLU, EU20-HellaSwag, EU20-ARC, EU20-TruthfulQA, and EU20-GSM8K \citep{thellmann2024multilingualllmevaluationeuropean}; or MMLU-Prox \citep{xuanMMLUProXMultilingualBenchmark2025b}. Other multilingual benchmarks were created with a special focus on cultural sensitivity by dividing the original subsets into culturally sensitive and culturally agnostic ones (Global MMLU, \citealp{singhGlobalMMLUUnderstanding2025}), or by using professional translators or multiple rounds of revision to raise the quality of the dataset, e.g., BenchMax \citep{huangBenchMAXComprehensiveMultilingual2025a}, Flores-101 and FLORES-200 \citep{goyal-etal-2022-flores} and Belebele \citep{bandarkar-etal-2024-belebele}. 

For speech, ML-SUPERB (Multilingual Speech processing Universal PERformance Benchmark, \citealp{shi2023mlsuperb}) extends the English SUPERB speech benchmark to 143 languages, evaluating self-supervised speech representations on automatic speech recognition and language identification. FLEURS~\cite{fleurs2022arxiv} is a speech-based extension of the FLORES multilingual benchmark, with focus on  language identification, automatic speech recognition, and retrieval evaluation. DIALECTBENCH \citep{faisal-etal-2024-dialectbench} is the first large-scale benchmark for language variety understanding, aggregating 10 text-level tasks for 281 varieties.

\subsection{Dynamic Benchmarks}
\label{dynamic}
Recent developments emphasize dynamic and contamination-resistant evaluation.
The period 2022--2025 has witnessed fundamental shifts toward more sophisticated evaluation approaches. One such attempt is LiveBench \citep{whiteLiveBenchChallengingContaminationLimited2025a}, the first benchmark designed to resist training data contamination through frequently updated questions from recent sources, automatic scoring, and monthly updates. This dynamic approach remains challenging, with the top models achieving accuracy below 80\%.

\subsection{Language- and Culture-Specific Benchmarks}
\label{culture-specific}
Focusing on different approaches to the evaluation of LLM performance in non-English European languages, we find a broad array of language- and culture-specific benchmarks developed with various methodologies and serving different purposes, of which we present the ones we find most interesting.

Many European languages have established evaluation frameworks dedicated to language-specific benchmarks, and in most cases such frameworks combine traditional datasets translated from English, and native more culture-aware benchmarks. Examples include HuLu\footnote{\url{https://hulu.nytud.hu}} for Hungarian \citep{ligeti-nagyHuLUHungarianLanguage2024}, which covers a number of well-known tasks such as plausible alternatives (HuCoPa), textual entailment (HuRTE) and linguistic acceptability (HuCoLa), of which the latter was originally constructed using sentences from selected Hungarian linguistics books. BenCzechMark for Czech \citep{fajcik2025benczechmarkczechcentricmultitask} is a complex benchmark collection comprising 50 tasks, of which 14 were newly created and only 10\% of the collective instances were machine-translated. The authors also employ multiple evaluation metrics including duel scoring. Another recent benchmark for Czech, Ukrainian and Slovak called CUS-QA \citep{libovický2025cusqalocalknowledgeorientedopenendedquestion} focuses specifically on cultural competence and crafts questions, both textual and visual, from Wikipedia articles which exist in only one of the languages. 

For Iberian languages, a comprehensive and extensible framework has been established under IberBench \citep{gonzález2025iberbenchllmevaluationiberian}, spanning 22 task categories and addressing both generic and industry-relevant tasks. In parallel and under a similar name, IberoBench \citep{baucells-etal-2025-iberobench} offers 62 tasks of which several were created from scratch from native data, and others were included only if they satisfied rather strict quality criteria. 

For Slovenian, the SloBENCH\footnote{\url{https://slobench.cjvt.si}} evaluation framework offers natural language inference (SI-NLI), machine translation, speech recognition, Slovene SuperGLUE, and two pragmatics benchmarks: SloPragMega and SloPragEval. To create the latter, full localization of the originally English dataset was performed, by adapting cultural references and occasionally completely rewriting examples to better match the linguistic and cultural context. A similar approach has been taken while translating and adapting the COPA benchmark~\citep{roemmele2011copa} to four standard languages and three dialects of the South Slavic language group, resulting in the DIALECT-COPA~\citep{ljubesic-etal-2024-dialect} benchmark collection. While proper cultural adaptation did take place, the overall popularity and age of the parent COPA benchmark surely makes it prone to LLM contamination. Regardless of that, today's best-performing proprietary models still score only halfway between random and optimal on dialectal data ~\citep{chifu-etal-2024-vardial}.

A fully native benchmark is ITALIC for Italian \citep{seveso-etal-2025-italic}, which comprises 10,000 instances from 12 domains and was built entirely from exam materials offered by various public institutions or government bodies. 

An exceptionally active approach to language- and culture-specific benchmarking can be observed for Polish\footnote{\url{https://huggingface.co/spaces/speakleash/polish-llm-benchmarks}}, with a range of generic and domain-specific evaluations for Polish including multi-turn conversation (MT-Bench), emotional intelligence (EQ-Bench), comprehensive text understanding (CPTUB\footnote{\url{https://huggingface.co/spaces/speakleash/cptu\_bench}}), medical domain benchmark, linguistic and cultural competency (PLCC, \citealp{dadasEvaluatingPolishLinguistic2025}), educational (LLMs Behind the School Desk), cultural vision benchmark, and legal QA tasks. Most of these benchmarks were developed anew, by carefully selecting tasks and examples, verifying them by experts and collecting human annotations. The CPTUB is composed of two parts, the first evaluating implicatures (implied meaning, sarcasm, idiomatic expressions) and the second testing on tricky questions (logical puzzles, semantic ambiguity, logical inconsistencies, absurdity, humor). Similarly, the PLCC consists of 600 manually crafted questions and is divided into six categories: history, geography, culture and tradition, art and entertainment, grammar, and vocabulary. The leaderboard results\footnote{\url{https://huggingface.co/spaces/sdadas/plcc}} indicate that even the largest models still reach mediocre performance in Polish grammar and vocabulary, thus justifying the need for detailed assessment of linguistic competence for other European languages as well. A final example of a culturally specific benchmark is the Polish Cultural Vision Benchmark\footnote{\url{https://huggingface.co/spaces/speakleash/Polish_Cultural_Vision_Benchmark}}, a collection of images with text descriptions to evaluate the cultural competence of multimodal models. The dataset is part of a citizen science project aimed at collecting 1 million culturally specific images\footnote{\url{https://obywatel.bielik.ai}} and recruiting user donations under the slogan of ``technopatriotism.'' While similar platforms have been established before to collect text data, this is a positive example of a contemporary and at the same time participatory benchmark. 

\section{Categorization of Benchmarks}
\label{taxonomy}
Since this paper makes a case for a European database of LLM benchmarks, we propose a new taxonomy which would allow a better categorization and labeling of benchmarks for non-English languages. This would allow us to better compare LLMs across languages; gain deeper insight into the strengths and weaknesses of current and future LLMs in a specific language, use case, modality or domain; set common priorities and work towards filling the evaluation and performance gaps.

The proposed taxonomy should serve as a (tentative) \textit{hierarchy of labels} to organize or classify benchmarks; many (or indeed most) belong to more than a single category. For this reason, a natural choice to store and query the European benchmarking activities is a database, in which benchmarks are described according to this proposal. Thus, the benchmarks are assigned non-exclusive categories and are richly described with metadescriptions.

While alignment, including trustworthiness, truthfulness and safety of LLMs, are central topics to the development of LLMs, they constitute another level of evaluation. Many elements of AI ethics partly overlap, or are entailed, in other benchmarks (e.g., bias is revealed in translation or language generation; trustworthiness is related to reasoning performance, etc.). This is another reason why multiple categories per benchmark are considered typical and expected.

Since LLMs tend to perform worse in non-English languages, and especially non-standard varieties, over a spectrum of tasks, we propose that the language- and culture-related abilities receive more attention, and therefore a more fine-grained taxonomy than they do in existing taxonomies. The list can be expanded as needed.

\subsection{Existing Taxonomies}
As new benchmarks are continuously presented to evaluate the emerging capabilities of LLMs, many attempts have been made to organize them in a structured and logical way. 

The \textbf{AI Verify Foundation} has established one of the most systematic approaches to LLM benchmark categorization globally. In their October 2023 publication ``Cataloguing LLM Evaluations'' \citep{aiverify_catalogue}, LLM benchmarks are organized into 5 top categories (further divided into subcategories). These are \textit{General Capabilities} (natural language understanding, natural language generation, reasoning, knowledge and factuality, tool use effectiveness, multilingualism, and context length handling); \textit{Domain Specific Capabilities} (specialized industry performance across various domains); \textit{Safety and Trustworthiness}; \textit{Extreme Risks}; and \textit{Undesirable Use Cases}. 

The catalogue represents a comprehensive and valuable contribution to the field, and has many positive features: The taxonomy is based on LLM capabilities, occasionally also referred to as tasks, which seems intuitively most pragmatic as this is usually the way we think about (and evaluate) human performance too. Complex benchmarks can appear in several categories simultaneously (e.g., BigBench as a massive collaborative benchmark appears in almost all taxonomy categories), and the recommendations for future LLM evaluations are a solid starting point to reinforce minimum quality standards for fair and trustworthy LLM assessment. 

However, the catalogue also has some drawbacks which render it unsuitable for our purposes. Firstly, to no fault of its authors, it has not been updated since 2023 and hence does not include many benchmarks which have since become mainstream, nor does it address recent developments in LLMs and AI in general. Secondly, although it includes Multilinguality as a separate category, it falls short in capturing some aspects of LLM performance which may be critical for the evaluation of European models; i.e., models specifically developed to be used in region-, language-, culture- or domain-specific contexts. Thirdly, and this is less of a drawback but simply an observation, the taxonomy and the quality recommendations are primarily focused on the safety and trustworthiness of LLMs, in the context of AI governance and alignment research. While these are indeed crucial priorities especially for the so-called ''frontier models'' and capabilities, the European landscape of LLM development and evaluation is -- at least for now -- gyrating around a different set of goals, such as how to reach state-of-the-art levels of understanding and generation in non-English languages, or how to de-bias English-centric models. 

Other approaches to taxonomization include HELM (Holistic Evaluation of Language Models), also referred to as the Stanford approach \citep{liang2023holisticevaluationlanguagemodels}. The authors introduce the concept of scenarios (what we want to evaluate) and metrics (which performance aspects are measured, and how), then propose a taxonomy of scenarios and desiderata. Today, the framework\footnote{\url{https://crfm.stanford.edu/helm/}} includes a number of leaderboards with support for multimodality and model-graded evaluation. While the scenarios proposed in HELM and the framework itself leave room for continuous extension, they do not in fact offer a hierarchical structure with sufficient focus on multilinguality and issues related to the use of LLMs in non-English contexts.

Similarly, \citet{chang2023surveyevaluationlargelanguage} provide an overview of existing LLM evaluations, which they examine from three aspects: what, where, and how to evaluate. They divide the evaluations tasks into eight top-level non-exclusive categories, namely \textit{Natural language processing}; \textit{Robustness, ethics, biases and trustworthiness}; \textit{Social sciences}; \textit{Natural science and engineering}; \textit{Medical applications}; \textit{Agent applications}; and \textit{Other applications}.

\citet{huberLLMsMeetBlooms} present a cognitive-based view on benchmarking by mapping the well-known Bloom's taxonomy of cognitive abilities to LLM capabilities across six hierarchical cognitive levels: \textit{Remember,} \textit{Understand}, \textit{Apply}, \textit{Analyze}, \textit{Evaluate}, and \textit{Create}, revealing significant gaps in the coverage of higher-order thinking skills.

Another comprehensive attempt at taxonomizing benchmarks is by \citet{guoEvaluatingLargeLanguage2023b} who introduce a three-pillar framework that categorizes LLM evaluation into three major groups: \textit{Knowledge and capability evaluation}, \textit{Alignment evaluation}, and \textit{Safety evaluation}.

\subsection{New Taxonomy Proposal}

As already outlined in the sections above, the proposed taxonomy, provided in Appendix \ref{sec:appendix}, is intended for the categorization of (mainly) European LLM benchmarks and is based on AI Verify Foundation's catalogue, but with the following modifications:
\begin{itemize}
\item We merge all \textbf{language-related tasks and scenarios} under a single top-level category called Language capabilities (see Table \ref{tab:llm_benchmarks} in Appendix). 
\item We further \textbf{merge the traditional NLP divide between natural language understanding and natural language generation} into a single subcategory. The fact is that state-of-the-art LLMs more often than not combine these two capabilities, and even straight-forward tasks such as question answering or text summarization involve both. 
\item We \textbf{expand the category for general linguistic competence} with further subcategories for style, conversation and pragmatics, and allow for other more fine-grained aspects of measuring the grammaticality, stylistic appropriateness or coherence of generated outputs (see Tables \ref{tab:llm_benchmarks} and \ref{tab:llm_benchmarks-part2} in Appendix).
\item We \textbf{expand the category of specific linguistic competence} to include creativity, atypical communication, the use of domain-specific language etc. (see Table \ref{tab:llm_benchmarks-part2}).
\item We also \textbf{expand and redefine the category of multilinguality} to include code-switching, multilingual interaction, and dialectal flexibility (see Table \ref{tab:llm_benchmarks-part2}). 
\item We introduce \textbf{cultural competence} as a separate category.
\item We introduce \textbf{speech} as a separate category to include benchmarks specifically aimed at performing tasks unique to spoken language as input or output (see Table \ref{tab:llm_benchmarks-part2}).\footnote{We propose Modality as one of the metadescriptors, allowing for any benchmark to be implemented in any of the modes. The Speech category refers to evaluations targeted at speech-related activities. }
\item We add \textbf{agency} as a form of long-term, consistent or strategic reasoning (see Table \ref{tab:llm_benchmarks_part3}). 

\end{itemize}

\begin{figure}[!ht]
\begin{center}
\includegraphics[width=\columnwidth]{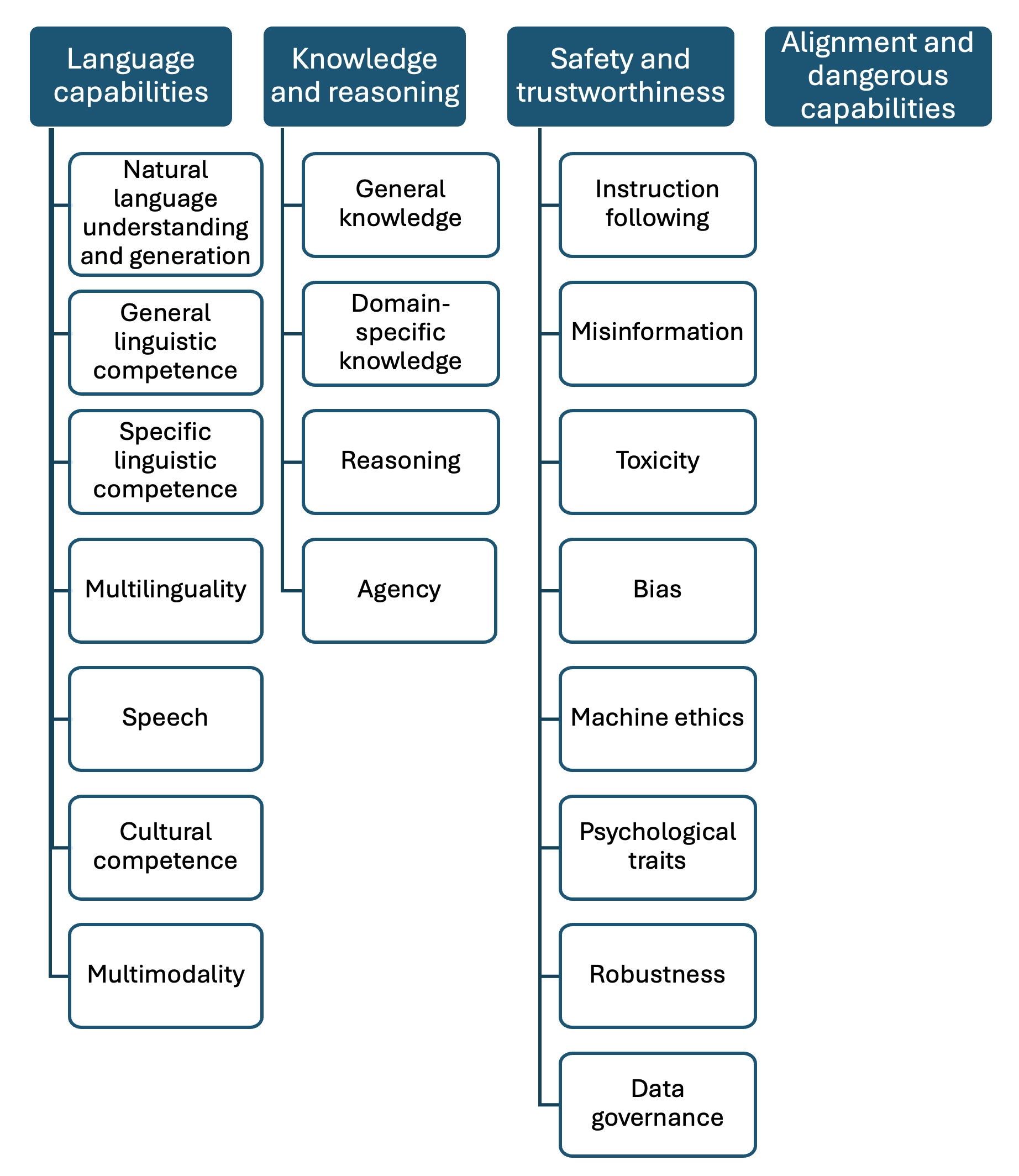}
\caption{Top-level categories with subcategories.}
\label{fig.1}
\end{center}
\end{figure}

Figure \ref{fig.1} shows the four top-level taxonomy categories with subcategories. The full taxonomy along with fine-grained third-level categories is provided in Tables \ref{tab:llm_benchmarks}, \ref{tab:llm_benchmarks-part2}, \ref{tab:llm_benchmarks_part3} and \ref{tab:llm_benchmarks_part4} in Appendix \ref{sec:appendix}.

\section{Quality Standards and Metadata}
\label{quality}

If we conceive of a European benchmarking registry as a searchable database which may help the LLM development community in setting priorities, exchanging knowledge and standardizing evaluation practices, we should be able to define some desiderata around how a particular benchmark is presented, described and distributed. Below we discuss some of these aspects.

\subsection{Provenance}

While it is clear that the development of original and sufficiently complex benchmarking datasets is highly time-consuming and costly, the drawbacks of automatically translated and culturally maladapted benchmarks have been clearly pointed out \citep{singhGlobalMMLUUnderstanding2025,xuanMMLUProXMultilingualBenchmark2025b}. We should thus strive towards clearer -- even if more complex -- descriptors which indicate how a dataset or benchmark was created. We propose the following descriptors:
\begin{itemize}
    \item \textbf{Original} Applies to datasets which have been originally created in the language they appear in, \textit{by any method other than translation} (e.g., collecting original exam questions, employing experts to provide domain-specific tasks, adapting authentic texts in a particular language to create tasks in that language).
\item \textbf{Machine translation} Applies to datasets created \textit{by any automated translation service}, including those created by LLMs using any kind of prompt scenario, and workflows with machine revision. The tools and workflows must be specified.
\item \textbf{Machine translation and Human revision} Applies to datasets where the result of \textit{automated translation was revised by a human} professional or non-professional translator or reviewer. Since many non-English benchmarks are created by machine translating the -- usually -- English original, followed by human revision of only a small portion of the dataset, the recommendation would be to use all labels that apply.
\item \textbf{Human translation} Used only for datasets which have been \textit{fully translated and revised by humans}. Only a few European non-English datasets satisfy this criterion.
\item \textbf{Full localization} Used for datasets which have not only been translated professionally, but for which a \textit{full linguistic and cultural adaptation} was performed. This might mean the replacement of culturally-specific or untranslatable tasks with new ones, or removing parts of the dataset deemed culturally unsuitable.
\item \textbf{Other} If several of the above scenarios were used, the dataset should be labeled with all that apply. Other methods and scenarios used in the creation of a dataset should be specified here.
\end{itemize}

\subsection{Accessibility}

The tension between open benchmarking and data contamination poses a significant challenge for AI evaluation. While public datasets enable reproducible research and fair comparisons, they risk contamination when models train on test data, inflating performance scores and undermining benchmark validity.
Private evaluation sets offer a potential solution by keeping data hidden from training processes, ensuring cleaner performance measurements. 
\begin{itemize}
    \item \textbf{Public} Applies to fully open datasets shared with labels through common platforms.
\item \textbf{Public without labels} Applies to datasets where labels are not distributed to prevent direct training.
\item \textbf{Private (academic/research access)} Where authors encourage reproducibility but wish to prevent contamination.
\item \textbf{Private (closed/proprietary)} Where datasets are typically not shared as they are used for internal or industry-specific evaluation.
\item \textbf{Other} This may include dynamic benchmarks where tasks are continually updated (such as \citealp{whiteLiveBenchChallengingContaminationLimited2025a}).
\end{itemize} 

\subsection{Language Coverage}

This category indicates the prominence, reach or scale of a benchmark in terms of its presence on major leaderboards, coverage, global spread, but also purpose. We are aware that the boundaries between the proposed buckets may be fuzzy.
\begin{itemize}
    \item \textbf{Major global benchmark} See section \ref{major-int} for examples.
    \item \textbf{Multilingual benchmark} This category can be used for benchmarks derived from the above, for instance by developing a multilingual variant of a well-known benchmark for a set of new languages. Examples include XCOPA \citep{ponti-etal-2020-xcopa}, MMLU-Prox \citep{xuanMMLUProXMultilingualBenchmark2025b} or xHumanEval \citep{raihan-etal-2025-mhumaneval}. 
    \item \textbf{Language-group or region-specific benchmark} This category is used for language-specific benchmarks as well as benchmarks that cover multiple languages from a similar language group or target a certain geographical region. Examples include IberoBENCH~\citep{baucells-etal-2025-iberobench} and DIALECT-COPA~\citep{ljubesic-etal-2024-dialect}. 
    \item \textbf{Other}
\end{itemize}

\subsection{Evaluation Type}
An important factor for present and future benchmarks is the divide between \textbf{closed-ended} types of tasks, most prominently multiple-choice questions, but also other types of tasks where the solution is included in the task, and \textbf{open-ended} tasks, typically generation of text, speech, image, or multimodal output. Few benchmarks to date address the latter, despite the fact that generative LLMs are now mainstream and the vast majority of application scenarios exploit generative abilities.

\subsection{Evaluation Metrics and Frameworks}

The performance of an LLM can be evaluated in several ways, depending on the type of task. For multiple choice questions, text classification tasks or cloze tests, where the correct answer is deterministic, \textbf{accuracy} or \textbf{F1} can be used. However, to evaluate longer, more complex responses resulting from generative tasks, many other methods were proposed. In reference-based evaluation, the LLM response is compared to a reference using various distance measures (e.g., \textbf{BERTScore, Rouge-1, METEOR}), while in reference-less contexts, the quality of the response is directly assessed (e.g., by an \textbf{LLM-as-a-judge}). As we have seen, recently developed benchmarks employ more complex evaluation methodologies, and a common alternative to algorithmic benchmarks is human preference voting (in so-called chatbot arenas, e.g., \url{https://lmarena.ai/}).

Another important element is the existence of a human baseline, and its quality. Important factors to consider are participant selection (demographic sampling, expertise), participant training, task design and instructions (to ensure fair comparison between humans and LLMs), control for attention, bias or fatigue, and the collection of participant demographics to facilitate reproducibility and interpretation of results. Human annotators or participants can also provide relevant insight into the overall quality of the benchmark, thus some campaigns actively collect human feedback to be used in revised versions of the evaluation dataset. 

If the benchmark or dataset is integrated into an evaluation framework, this should be indicated together with a link or other reference to the evaluation site.

\subsection{Metadescriptors}

We propose collecting rich metadata for each benchmark that allows researchers to quickly understand its content, characteristics, and provenance. Such metadescriptors facilitate dataset discovery, comparison, and reuse. Below, we summarize the metadata fields currently envisaged in the proposed registry:
\begin{itemize}
    \item \textbf{Description}: A short summary of the dataset’s content and purpose.
    \item \textbf{Benchmark family}: The broader benchmark initiative or collection to which the dataset belongs. For example, the COPA benchmark family would include the English COPA \citep{roemmele2011copa} and its many parallel variants such as the COPA datasets in Hungarian \citep{ligeti-nagyHuLUHungarianLanguage2024}, Croatian (\citealp{ljubevsic2021copa-hr}), South Slavic dialects \citep{ljubesic-etal-2024-dialect} etc.
    \item \textbf{Number of test instances}: The total number of instances in the test set, enabling quick comparison of dataset scale.
    \item \textbf{Language}: The language(s) in which the dataset is provided.
    \item \textbf{Language type}: Specification of whether the language of the dataset is standard, non-standard, or a dialect.
    \item \textbf{Modality}: The input modality of the dataset, such as text, speech, sign language, or audiovisual signal.
    \item \textbf{Authors}: The creators or curators of the dataset.
    \item \textbf{Paper link}: Reference to the main publication describing the dataset.
    \item \textbf{Access info}: Information on how to obtain the dataset, e.g., a link to a website from where the dataset can be acquired, or an e-mail of the dataset owner if the dataset is not published online.
    \item \textbf{Last revised}: The date of the most recent update or revision of the dataset.
    \item \textbf{More information}: Additional notes, links, or resources relevant to the dataset.
\end{itemize}

\section{Recommendations and Future Directions}

Several challenges of LLM evaluation have been pointed out by a number of studies (e.g., \citealp{laskar-etal-2024-systematic} or \citealp{aiverify_catalogue}: p. 16--22), most notably reproducibility, reliability (including contamination, obscure evaluation methods and unfair comparisons), and robustness. 
There are numerous parallel activities in progress to set the course of the European LLM evaluation landscape, agree on common principles and establish a dialogue between different stakeholders. While a full review of the above-mentioned challenges is beyond the scope of this paper, we list some priorities which apply in particular for the European benchmarking landscape and evaluations in non-English settings.
\begin{enumerate}
    
    \item \textbf{Cultural sensitivity} The prevalent framing of toxicity, bias, but also factual knowledge and reasoning, tends to be Western-centric. Evaluation concepts should be expanded to include diverse global perspectives and values.
      
    \item \textbf{Language sensitivity} Under this term, we refer to the fact that LLMs find applications in multilingual and multicultural settings, and that the performance of an LLM may depend heavily on the language it is used in. Benchmarks for bias and toxicity, but also some common benchmarks for language understanding and inference, are not trivial to transfer across languages and linguistic structures.
    
    \item \textbf{Comparability vs. specificity} Many European and other non-English languages still have a limited number of benchmark datasets. To provide a common ground for the evaluation of LLMs across languages, major global benchmarks are still being translated to facilitate international comparability. It is recommended that such translation and adaptation procedures acknowledge the complexity of the task and, even if performed automatically, consider employing different tools, prompt strategies, LLM revision techniques and human evaluation of random samples. 
    \item \textbf{Additional modalities} Many benchmarks have been developed recently to include different modalities, but there is still a gap in benchmarks for speech, sign language, audiovisual communication, and combinations thereof. 
    
    \item \textbf{Human baseline} While frontier LLMs already surpass humans in many areas, it is of paramount importance, especially in high-risk domains, to have a solid foundation for the evaluation of LLMs by comparing the results to human performance on a specific task, acknowledging that every benchmark is a limited representation of a task.
    \item \textbf{Transparent implementation} The details of the evaluation process should be published in a transparent manner, including the exact specification of prompting strategies and hyperparameters used, such as temperature. If possible, evaluation scripts should also be made public, and results validated through multiple runs in the case of smaller datasets.
    \item \textbf{Context-specific evaluation} There is a lack of nuanced, context-specific evaluations that address the multi-faceted nature of real-world LLM deployments. This includes domain-specificity, but also other elements of attuning evaluation to the users it serves.
\end{enumerate}
    
We believe that the rapidly evolving benchmarking landscape for European languages would benefit from a catalogue where each benchmark is categorized and documented according to the above principles. We conceive of such a catalogue in the form of a database, accommodating the fact that the proposed taxonomy categories and metadescriptors may overlap, so that each entry can be labeled with all which apply.

\section{Conclusion}

We have presented recent trends in LLM benchmarking for European languages and proposed a new taxonomy for their categorization, intended to be implemented alongside a range of metadescriptors in the context of a European catalogue of LLM benchmarks. Our taxonomization strategy focuses on the linguistic, cultural, factual and reasoning capabilities of models and also incorporates emerging abilities. The proposed considerations follow the wide-spread belief amongst European researchers and developers that the traditional Western-centric, likely contaminated and linguistically inappropriate datasets no longer satisfy our needs, and that targeted efforts should be invested in filling the evaluation gaps for all European languages. 

The initiative presented in this paper is the result of a series of discussions and reflections within the framework of several international research communities, collecting and integrating feedback from a number of researchers, developers and benchmark creators. With the rapid advancement of the field, we envisage continuous extensions and revisions both of the taxonomy and the associated set of metadescriptors and recommendations.

\section*{Acknowledgements}

This work is supported by 
the LLM4DH project, funded by the Slovenian Research and Innovation Agency (ARIS) under the grant agreement GC-0002.

\section{Bibliographical References}
\label{sec:reference}
\urlstyle{same}
\bibliographystyle{lrec2026-natbib}
\bibliography{LREC2026-Charting-benchmarks/benchmarking}

\begin{thebibliography}{71}
\expandafter\ifx\csname natexlab\endcsname\relax\def\natexlab#1{#1}\fi

\bibitem[{{AI Verify Foundation}(2023)}]{aiverify_catalogue}
{AI Verify Foundation}. 2023.
\newblock \href {https://aiverifyfoundation.sg/wp-content/uploads/2024/05/Cataloguing_LLM_Evaluations.pdf} {{Catalogue of LLM Evaluations}}.

\bibitem[{Bandarkar et~al.(2024)Bandarkar, Liang, Muller, Artetxe, Shukla, Husa, Goyal, Krishnan, Zettlemoyer, and Khabsa}]{bandarkar-etal-2024-belebele}
Lucas Bandarkar, Davis Liang, Benjamin Muller, Mikel Artetxe, Satya~Narayan Shukla, Donald Husa, Naman Goyal, Abhinandan Krishnan, Luke Zettlemoyer, and Madian Khabsa. 2024.
\newblock \href {https://doi.org/10.18653/v1/2024.acl-long.44} {The {Belebele} {Benchmark}: a {Parallel} {Reading} {Comprehension} {Dataset} in 122 {Language} {Variants}}.
\newblock In \emph{Proceedings of the 62nd {Annual} {Meeting} of the {Association} for {Computational} {Linguistics} ({Volume} 1: {Long} {Papers})}, pages 749--775, Bangkok, Thailand. Association for Computational Linguistics.

\bibitem[{Baucells et~al.(2025)Baucells, Aula-Blasco, de~Dios-Flores, Paniagua~Su{\'a}rez, Perez, Salles, Sotelo~Docio, Falc{\~a}o, Saiz, Sepulveda~Torres, Barnes, Gamallo, Gonzalez-Agirre, Rigau, and Villegas}]{baucells-etal-2025-iberobench}
Irene Baucells, Javier Aula-Blasco, Iria de~Dios-Flores, Silvia Paniagua~Su{\'a}rez, Naiara Perez, Anna Salles, Susana Sotelo~Docio, J{\'u}lia Falc{\~a}o, Jose~Javier Saiz, Robiert Sepulveda~Torres, Jeremy Barnes, Pablo Gamallo, Aitor Gonzalez-Agirre, German Rigau, and Marta Villegas. 2025.
\newblock \href {https://aclanthology.org/2025.coling-main.699/} {{I}bero{B}ench: A benchmark for {LLM} evaluation in {I}berian languages}.
\newblock In \emph{Proceedings of the 31st International Conference on Computational Linguistics}, pages 10491--10519, Abu Dhabi, UAE. Association for Computational Linguistics.

\bibitem[{Borkan et~al.(2019)Borkan, Dixon, Sorensen, Thain, and Vasserman}]{borkan2019nuanced}
Daniel Borkan, Lucas Dixon, Jeffrey Sorensen, Nithum Thain, and Lucy Vasserman. 2019.
\newblock \href {https://dl.acm.org/doi/abs/10.1145/3308560.3317593} {Nuanced metrics for measuring unintended bias with real data for text classification}.
\newblock In \emph{Companion proceedings of the 2019 world wide web conference}, pages 491--500.

\bibitem[{Chang et~al.(2023)Chang, Wang, Wang, Wu, Yang, Zhu, Chen, Yi, Wang, Wang, Ye, Zhang, Chang, Yu, Yang, and Xie}]{chang2023surveyevaluationlargelanguage}
Yupeng Chang, Xu~Wang, Jindong Wang, Yuan Wu, Linyi Yang, Kaijie Zhu, Hao Chen, Xiaoyuan Yi, Cunxiang Wang, Yidong Wang, Wei Ye, Yue Zhang, Yi~Chang, Philip~S. Yu, Qiang Yang, and Xing Xie. 2023.
\newblock \href {http://arxiv.org/abs/2307.03109} {A survey on evaluation of large language models}.
\newblock \emph{arXiv preprint arXiv:2307.03109}.

\bibitem[{Chifu et~al.(2024)Chifu, Glava{\v{s}}, Ionescu, Ljube{\v{s}}i{\'c}, Mileti{\'c}, Mileti{\'c}, Scherrer, and Vuli{\'c}}]{chifu-etal-2024-vardial}
Adrian-Gabriel Chifu, Goran Glava{\v{s}}, Radu~Tudor Ionescu, Nikola Ljube{\v{s}}i{\'c}, Aleksandra Mileti{\'c}, Filip Mileti{\'c}, Yves Scherrer, and Ivan Vuli{\'c}. 2024.
\newblock \href {https://doi.org/10.18653/v1/2024.vardial-1.1} {{V}ar{D}ial evaluation campaign 2024: Commonsense reasoning in dialects and multi-label similar language identification}.
\newblock In \emph{Proceedings of the Eleventh Workshop on NLP for Similar Languages, Varieties, and Dialects (VarDial 2024)}, pages 1--15, Mexico City, Mexico. Association for Computational Linguistics.

\bibitem[{Chizhov et~al.(2025)Chizhov, Nee, Langlais, and Yamshchikov}]{chizhovWhatHellaSwagValidity2025}
Pavel Chizhov, Mattia Nee, Pierre-Carl Langlais, and Ivan~P Yamshchikov. 2025.
\newblock \href {http://arxiv.org/abs/2504.07825} {{What the HellaSwag? On the Validity of Common-Sense Reasoning Benchmarks}}.
\newblock \emph{arXiv preprint arXiv:2504.07825}.

\bibitem[{Clark et~al.(2019)Clark, Lee, Chang, Kwiatkowski, Collins, and Toutanova}]{clark-etal-2019-boolq}
Christopher Clark, Kenton Lee, Ming-Wei Chang, Tom Kwiatkowski, Michael Collins, and Kristina Toutanova. 2019.
\newblock \href {https://doi.org/10.18653/v1/N19-1300} {{{B}ool{Q}: Exploring the Surprising Difficulty of Natural Yes/No Questions}}.
\newblock In \emph{Proceedings of the 2019 Conference of the North {A}merican Chapter of the Association for Computational Linguistics: Human Language Technologies, Volume 1 (Long and Short Papers)}, pages 2924--2936, Minneapolis, Minnesota. Association for Computational Linguistics.

\bibitem[{Clark et~al.(2018)Clark, Cowhey, Etzioni, Khot, Sabharwal, Schoenick, and Tafjord}]{clark2018thinksolvedquestionanswering}
Peter Clark, Isaac Cowhey, Oren Etzioni, Tushar Khot, Ashish Sabharwal, Carissa Schoenick, and Oyvind Tafjord. 2018.
\newblock \href {https://arxiv.org/abs/1803.05457} {{Think you have Solved Question Answering? Try ARC, the AI2 Reasoning Challenge}}.
\newblock \emph{arXiv preprint arXiv:1803.05457}.

\bibitem[{Conneau et~al.(2022)Conneau, Ma, Khanuja, Zhang, Axelrod, Dalmia, Riesa, Rivera, and Bapna}]{fleurs2022arxiv}
Alexis Conneau, Min Ma, Simran Khanuja, Yu~Zhang, Vera Axelrod, Siddharth Dalmia, Jason Riesa, Clara Rivera, and Ankur Bapna. 2022.
\newblock \href {https://arxiv.org/abs/2205.12446} {Fleurs: Few-shot learning evaluation of universal representations of speech}.
\newblock \emph{arXiv preprint arXiv:2205.12446}.

\bibitem[{Dadas et~al.(2025)Dadas, Gr{\k{e}}bowiec, Pere{\l}kiewicz, and Po{\'s}wiata}]{dadasEvaluatingPolishLinguistic2025}
S{\l}awomir Dadas, Ma{\l}gorzata Gr{\k{e}}bowiec, Micha{\l} Pere{\l}kiewicz, and Rafa{\l} Po{\'s}wiata. 2025.
\newblock \href {http://arxiv.org/abs/2503.00995} {{Evaluating Polish linguistic and cultural competency in large language models}}.
\newblock \emph{arXiv preprint arXiv:2503.00995}.

\bibitem[{Danescu-Niculescu-Mizil et~al.(2013)Danescu-Niculescu-Mizil, Sudhof, Jurafsky, Leskovec, and Potts}]{danescu2013computational}
Cristian Danescu-Niculescu-Mizil, Moritz Sudhof, Dan Jurafsky, Jure Leskovec, and Christopher Potts. 2013.
\newblock \href {https://aclanthology.org/P13-1025.pdf} {A computational approach to politeness with application to social factors}.
\newblock In \emph{Proceedings of the 51st Annual Meeting of the Association for Computational Linguistics (Volume 1: Long Papers)}, pages 250--259.

\bibitem[{De~Marneffe et~al.(2019)De~Marneffe, Simons, and Tonhauser}]{de2019commitmentbank}
Marie-Catherine De~Marneffe, Mandy Simons, and Judith Tonhauser. 2019.
\newblock \href {https://ojs.ub.uni-konstanz.de/sub/index.php/sub/article/view/601} {{The CommitmentBank: Investigating projection in naturally occurring discourse}}.
\newblock In \emph{Proceedings of Sinn und Bedeutung}, volume~23, pages 107--124.

\bibitem[{Etxaniz et~al.(2024)Etxaniz, Azkune, Soroa, Lacalle, and Artetxe}]{etxaniz2024bertaqalanguagemodelsknow}
Julen Etxaniz, Gorka Azkune, Aitor Soroa, Oier Lacalle, and Mikel Artetxe. 2024.
\newblock \href {https://proceedings.neurips.cc/paper_files/paper/2024/hash/3bb42f6bb1b1ab6809afd6c90865b087-Abstract-Datasets_and_Benchmarks_Track.html} {{BertaQA: How Much Do Language Models Know About Local Culture?}}
\newblock \emph{Advances in Neural Information Processing Systems}, 37:34077--34097.

\bibitem[{Faisal et~al.(2024)Faisal, Ahia, Srivastava, Ahuja, Chiang, Tsvetkov, and Anastasopoulos}]{faisal-etal-2024-dialectbench}
Fahim Faisal, Orevaoghene Ahia, Aarohi Srivastava, Kabir Ahuja, David Chiang, Yulia Tsvetkov, and Antonios Anastasopoulos. 2024.
\newblock \href {https://doi.org/10.18653/v1/2024.acl-long.777} {{DIALECTBENCH}: An {NLP} benchmark for dialects, varieties, and closely-related languages}.
\newblock In \emph{Proceedings of the 62nd Annual Meeting of the Association for Computational Linguistics (Volume 1: Long Papers)}, pages 14412--14454, Bangkok, Thailand. Association for Computational Linguistics.

\bibitem[{Fajcik et~al.(2025)Fajcik, Docekal, Dolezal, Ondrej, Bene{\v{s}}, Kapsa, Smrz, Polok, Hradis, Neverilova et~al.}]{fajcik2025benczechmarkczechcentricmultitask}
Martin Fajcik, Martin Docekal, Jan Dolezal, Karel Ondrej, Karel Bene{\v{s}}, Jan Kapsa, Pavel Smrz, Alexander Polok, Michal Hradis, Zuzana Neverilova, et~al. 2025.
\newblock \href {https://direct.mit.edu/tacl/article/doi/10.1162/TACL.a.32/132962} {{BenCzechMark : A Czech-centric Multitask and Multimetric Benchmark for Large Language Models with Duel Scoring Mechanism}}.
\newblock \emph{Transactions of the Association for Computational Linguistics}, 13:1068--1095.

\bibitem[{Gao et~al.(2024)Gao, Song, Yang, Cai, Miao, Dong, Li, Ma, Chen, Xu et~al.}]{gao2024omni}
Bofei Gao, Feifan Song, Zhe Yang, Zefan Cai, Yibo Miao, Qingxiu Dong, Lei Li, Chenghao Ma, Liang Chen, Runxin Xu, et~al. 2024.
\newblock \href {https://arxiv.org/abs/2410.07985} {Omni-math: A universal olympiad level mathematic benchmark for large language models}.
\newblock \emph{arXiv preprint arXiv:2410.07985}.

\bibitem[{Giampiccolo et~al.(2007)Giampiccolo, Magnini, Dagan, and Dolan}]{giampiccolo2007third-rte}
Danilo Giampiccolo, Bernardo Magnini, Ido Dagan, and William~B Dolan. 2007.
\newblock \href {https://aclanthology.org/W07-1401.pdf} {The third pascal recognizing textual entailment challenge}.
\newblock In \emph{Proceedings of the ACL-PASCAL workshop on textual entailment and paraphrasing}, pages 1--9.

\bibitem[{Gonz{\'a}lez et~al.(2025)Gonz{\'a}lez, Obrador, Herrero, Sarvazyan, Chinea-R{\'\i}os, Basile, and Franco-Salvador}]{gonzález2025iberbenchllmevaluationiberian}
Jos{\'e}~{\'A}ngel Gonz{\'a}lez, Ian~Borrego Obrador, {\'A}lvaro~Romo Herrero, Areg~Mikael Sarvazyan, Mara Chinea-R{\'\i}os, Angelo Basile, and Marc Franco-Salvador. 2025.
\newblock \href {https://arxiv.org/abs/2504.16921} {{IberBench: LLM Evaluation on Iberian Languages}}.
\newblock \emph{arXiv preprint arXiv:2504.16921}.

\bibitem[{Goyal et~al.(2022)Goyal, Gao, Chaudhary, Chen, Wenzek, Ju, Krishnan, Ranzato, Guzm{\'a}n, and Fan}]{goyal-etal-2022-flores}
Naman Goyal, Cynthia Gao, Vishrav Chaudhary, Peng-Jen Chen, Guillaume Wenzek, Da~Ju, Sanjana Krishnan, Marc{'}Aurelio Ranzato, Francisco Guzm{\'a}n, and Angela Fan. 2022.
\newblock \href {https://doi.org/10.1162/tacl_a_00474} {The {F}lores-101 evaluation benchmark for low-resource and multilingual machine translation}.
\newblock \emph{Transactions of the Association for Computational Linguistics}, 10:522--538.

\bibitem[{Guha et~al.(2023)Guha, Nyarko, Ho, R{\'e}, Chilton, Chohlas-Wood, Peters, Waldon, Rockmore, Zambrano et~al.}]{guha2023legalbench}
Neel Guha, Julian Nyarko, Daniel Ho, Christopher R{\'e}, Adam Chilton, Alex Chohlas-Wood, Austin Peters, Brandon Waldon, Daniel Rockmore, Diego Zambrano, et~al. 2023.
\newblock \href {https://proceedings.neurips.cc/paper_files/paper/2023/hash/89e44582fd28ddfea1ea4dcb0ebbf4b0-Abstract-Datasets_and_Benchmarks.html} {Legalbench: A collaboratively built benchmark for measuring legal reasoning in large language models}.
\newblock \emph{Advances in neural information processing systems}, 36:44123--44279.

\bibitem[{Guo et~al.(2023)Guo, Jin, Liu, Huang, Shi, Yu, Liu, Li, Xiong, Xiong et~al.}]{guoEvaluatingLargeLanguage2023b}
Zishan Guo, Renren Jin, Chuang Liu, Yufei Huang, Dan Shi, Linhao Yu, Yan Liu, Jiaxuan Li, Bojian Xiong, Deyi Xiong, et~al. 2023.
\newblock \href {http://arxiv.org/abs/2310.19736} {Evaluating {{Large Language Models}}: {{A Comprehensive Survey}}}.
\newblock \emph{arXiv preprint arXiv:2310.19736}.

\bibitem[{Hendrycks et~al.(2021)Hendrycks, Burns, Basart, Zou, Mazeika, Song, and Steinhardt}]{hendryckstest2021}
Dan Hendrycks, Collin Burns, Steven Basart, Andy Zou, Mantas Mazeika, Dawn Song, and Jacob Steinhardt. 2021.
\newblock \href {https://openreview.net/forum?id=d7KBjmI3GmQ} {{Measuring Massive Multitask Language Understanding}}.
\newblock \emph{Proceedings of the International Conference on Learning Representations (ICLR)}.

\bibitem[{Huang et~al.(2025)Huang, Zhu, Hu, He, Li, Huang, and Yuan}]{huangBenchMAXComprehensiveMultilingual2025a}
Xu~Huang, Wenhao Zhu, Hanxu Hu, Conghui He, Lei Li, Shujian Huang, and Fei Yuan. 2025.
\newblock \href {http://arxiv.org/abs/2502.07346} {{BenchMAX}: {A} {Comprehensive} {Multilingual} {Evaluation} {Suite} for {Large} {Language} {Models}}.
\newblock \emph{arXiv preprint arXiv:2502.07346}.

\bibitem[{Huber and Niklaus(2025)}]{huberLLMsMeetBlooms}
Thomas Huber and Christina Niklaus. 2025.
\newblock \href {https://aclanthology.org/2025.coling-main.350/} {{LLMs Meet Bloom's Taxonomy: A Cognitive View on Large Language Model Evaluations}}.
\newblock In \emph{Proceedings of the 31st International Conference on Computational Linguistics}, pages 5211--5246.

\bibitem[{Jin et~al.(2021)Jin, Pan, Oufattole, Weng, Fang, and Szolovits}]{jin2021disease}
Di~Jin, Eileen Pan, Nassim Oufattole, Wei-Hung Weng, Hanyi Fang, and Peter Szolovits. 2021.
\newblock \href {https://www.mdpi.com/2076-3417/11/14/6421} {What disease does this patient have? a large-scale open domain question answering dataset from medical exams}.
\newblock \emph{Applied Sciences}, 11(14):6421.

\bibitem[{Khashabi et~al.(2018)Khashabi, Chaturvedi, Roth, Upadhyay, and Roth}]{MultiRC2018}
Daniel Khashabi, Snigdha Chaturvedi, Michael Roth, Shyam Upadhyay, and Dan Roth. 2018.
\newblock \href {https://aclanthology.org/N18-1023/} {{Looking Beyond the Surface: A Challenge Set for Reading Comprehension over Multiple Sentences}}.
\newblock In \emph{Proceedings of North American Chapter of the Association for Computational Linguistics (NAACL)}.

\bibitem[{Klemen et~al.(2022)Klemen, {\v Z}agar, {\v C}ibej, and Robnik-{\v S}ikonja}]{si-nli-clarin}
Matej Klemen, Ale{\v s} {\v Z}agar, Jaka {\v C}ibej, and Marko Robnik-{\v S}ikonja. 2022.
\newblock \href {http://hdl.handle.net/11356/1707} {Slovene natural language inference dataset {SI}-{NLI}}.
\newblock Slovenian language resource repository {CLARIN}.{SI}.

\bibitem[{Kornilova and Eidelman(2019)}]{kornilova-eidelman-2019-billsum}
Anastassia Kornilova and Vladimir Eidelman. 2019.
\newblock \href {https://doi.org/10.18653/v1/D19-5406} {{B}ill{S}um: A corpus for automatic summarization of {US} legislation}.
\newblock In \emph{Proceedings of the 2nd Workshop on New Frontiers in Summarization}, pages 48--56, Hong Kong, China. Association for Computational Linguistics.

\bibitem[{Kuzman et~al.(2021)Kuzman, Brglez, Rupnik, and Ljube{\v s}i{\'c}}]{ginco-clarin}
Taja Kuzman, Mojca Brglez, Peter Rupnik, and Nikola Ljube{\v s}i{\'c}. 2021.
\newblock \href {http://hdl.handle.net/11356/1467} {Slovene web genre identification corpus {GINCO} 1.0}.
\newblock Slovenian language resource repository {CLARIN}.{SI}.

\bibitem[{Kuzman et~al.(2023)Kuzman, Mozeti{\v{c}}, and Ljube{\v{s}}i{\'c}}]{kuzman2023automatic}
Taja Kuzman, Igor Mozeti{\v{c}}, and Nikola Ljube{\v{s}}i{\'c}. 2023.
\newblock \href {https://www.mdpi.com/2504-4990/5/3/59} {Automatic genre identification for robust enrichment of massive text collections: Investigation of classification methods in the era of large language models}.
\newblock \emph{Machine Learning and Knowledge Extraction}, 5(3):1149--1175.

\bibitem[{Laskar et~al.(2024)Laskar, Alqahtani, Bari, Rahman, Khan, Khan, Jahan, Bhuiyan, Tan, Parvez, Hoque, Joty, and Huang}]{laskar-etal-2024-systematic}
Md~Tahmid~Rahman Laskar, Sawsan Alqahtani, M~Saiful Bari, Mizanur Rahman, Mohammad Abdullah~Matin Khan, Haidar Khan, Israt Jahan, Amran Bhuiyan, Chee~Wei Tan, Md~Rizwan Parvez, Enamul Hoque, Shafiq Joty, and Jimmy Huang. 2024.
\newblock \href {https://doi.org/10.18653/v1/2024.emnlp-main.764} {A systematic survey and critical review on evaluating large language models: Challenges, limitations, and recommendations}.
\newblock In \emph{Proceedings of the 2024 Conference on Empirical Methods in Natural Language Processing}, pages 13785--13816, Miami, Florida, USA. Association for Computational Linguistics.

\bibitem[{Levesque et~al.(2012)Levesque, Davis, and Morgenstern}]{levesque2012winograd}
Hector~J Levesque, Ernest Davis, and Leora Morgenstern. 2012.
\newblock \href {https://cdn.aaai.org/ocs/4492/4492-21843-1-PB.pdf} {{The Winograd Schema Challenge}}.
\newblock In \emph{13th International Conference on the Principles of Knowledge Representation and Reasoning, KR 2012}, pages 552--561. Institute of Electrical and Electronics Engineers Inc.

\bibitem[{Liang et~al.(2022)Liang, Bommasani, Lee, Tsipras, Soylu, Yasunaga, Zhang, Narayanan, Wu, Kumar et~al.}]{liang2023holisticevaluationlanguagemodels}
Percy Liang, Rishi Bommasani, Tony Lee, Dimitris Tsipras, Dilara Soylu, Michihiro Yasunaga, Yian Zhang, Deepak Narayanan, Yuhuai Wu, Ananya Kumar, et~al. 2022.
\newblock \href {https://arxiv.org/abs/2211.09110} {{Holistic Evaluation of Language Models}}.
\newblock \emph{arXiv preprint arXiv:2211.09110}.

\bibitem[{Libovick{\`y} et~al.(2025)Libovick{\`y}, Helcl, Manea, and Vico}]{libovický2025cusqalocalknowledgeorientedopenendedquestion}
Jind{\v{r}}ich Libovick{\`y}, Jind{\v{r}}ich Helcl, Andrei Manea, and Gianluca Vico. 2025.
\newblock \href {https://arxiv.org/abs/2507.22752} {{CUS-QA: Local-Knowledge-Oriented Open-Ended Question Answering Dataset}}.
\newblock \emph{arXiv preprint arXiv:2507.22752}.

\bibitem[{Ligeti-Nagy et~al.(2024)Ligeti-Nagy, Ferenczi, H{\'e}ja, Laki, Vad{\'a}sz, Yang, and V{\'a}radi}]{ligeti-nagyHuLUHungarianLanguage2024}
No{\'e}mi Ligeti-Nagy, Gerg{\H{o}} Ferenczi, Enik{\H{o}} H{\'e}ja, L{\'a}szl{\'o}~J{\'a}nos Laki, No{\'e}mi Vad{\'a}sz, Zijian~Gy{\H{o}}z{\H{o}} Yang, and Tam{\'a}s V{\'a}radi. 2024.
\newblock \href {https://aclanthology.org/2024.lrec-main.733/} {{HuLU: Hungarian language understanding benchmark kit}}.
\newblock In \emph{Proceedings of the 2024 Joint International Conference on Computational Linguistics, Language Resources and Evaluation (LREC-COLING 2024)}, pages 8360--8371.

\bibitem[{Lin et~al.(2022)Lin, Hilton, and Evans}]{lin2022truthfulqa}
Stephanie Lin, Jacob Hilton, and Owain Evans. 2022.
\newblock \href {https://aclanthology.org/2022.acl-long.229/} {{TruthfulQA: Measuring How Models Mimic Human Falsehoods}}.
\newblock In \emph{Proceedings of the 60th Annual Meeting of the Association for Computational Linguistics (Volume 1: Long Papers)}, pages 3214--3252.

\bibitem[{Ljube{\v s}i{\'c}(2021)}]{copa-hr-clarin}
Nikola Ljube{\v s}i{\'c}. 2021.
\newblock \href {http://hdl.handle.net/11356/1404} {{Choice of plausible alternatives dataset in Croatian {COPA}-{HR}}}.
\newblock Slovenian language resource repository {CLARIN}.{SI}.

\bibitem[{Ljube{\v{s}}i{\'c} et~al.(2024)Ljube{\v{s}}i{\'c}, Galant, Ben{\v{c}}ina, {\v{C}}ibej, Milosavljevi{\'c}, Rupnik, and Kuzman}]{ljubesic-etal-2024-dialect}
Nikola Ljube{\v{s}}i{\'c}, Nada Galant, Sonja Ben{\v{c}}ina, Jaka {\v{C}}ibej, Stefan Milosavljevi{\'c}, Peter Rupnik, and Taja Kuzman. 2024.
\newblock \href {https://doi.org/10.18653/v1/2024.vardial-1.7} {{DIALECT}-{COPA}: Extending the standard translations of the {COPA} causal commonsense reasoning dataset to {S}outh {S}lavic dialects}.
\newblock In \emph{Proceedings of the Eleventh Workshop on NLP for Similar Languages, Varieties, and Dialects (VarDial 2024)}, pages 89--98, Mexico City, Mexico. Association for Computational Linguistics.

\bibitem[{Ljube{\v{s}}i{\'c} and Lauc(2021)}]{ljubevsic2021copa-hr}
Nikola Ljube{\v{s}}i{\'c} and Davor Lauc. 2021.
\newblock \href {https://aclanthology.org/2021.bsnlp-1.5/} {{BERTi{\'c}-The Transformer Language Model for Bosnian, Croatian, Montenegrin and Serbian}}.
\newblock In \emph{Proceedings of the 8th Workshop on Balto-Slavic Natural Language Processing}, pages 37--42.

\bibitem[{Maas et~al.(2011)Maas, Daly, Pham, Huang, Ng, and Potts}]{maas-etal-2011-learning}
Andrew~L. Maas, Raymond~E. Daly, Peter~T. Pham, Dan Huang, Andrew~Y. Ng, and Christopher Potts. 2011.
\newblock \href {https://aclanthology.org/P11-1015/} {Learning word vectors for sentiment analysis}.
\newblock In \emph{Proceedings of the 49th Annual Meeting of the Association for Computational Linguistics: Human Language Technologies}, pages 142--150, Portland, Oregon, USA. Association for Computational Linguistics.

\bibitem[{Mochtak et~al.(2024)Mochtak, Rupnik, and Ljube{\v{s}}i{\'c}}]{mochtak2024parlasent}
Michal Mochtak, Peter Rupnik, and Nikola Ljube{\v{s}}i{\'c}. 2024.
\newblock \href {http://www.lrec-conf.org/proceedings/lrec-coling-2024/pdf/2024.main-1.1393.pdf} {The parlasent multilingual training dataset for sentiment identification in parliamentary proceedings}.
\newblock In \emph{Proceedings of the 2024 Joint International Conference on Computational Linguistics, Language Resources and Evaluation (LREC-COLING 2024)}, pages 16024--16036.

\bibitem[{Mochtak et~al.(2023)Mochtak, Rupnik, Meden, and Ljube{\v s}i{\'c}}]{parlasent-clarin}
Michal Mochtak, Peter Rupnik, Katja Meden, and Nikola Ljube{\v s}i{\'c}. 2023.
\newblock \href {http://hdl.handle.net/11356/1868} {The multilingual sentiment dataset of parliamentary debates {ParlaSent} 1.0}.
\newblock Slovenian language resource repository {CLARIN}.{SI}.

\bibitem[{Nallapati et~al.(2016)Nallapati, Zhou, dos Santos, Gu{\ensuremath{\dot{}}}l{\c{c}}ehre, and Xiang}]{nallapati-etal-2016-abstractive}
Ramesh Nallapati, Bowen Zhou, Cicero dos Santos, {\c{C}}a{\u{g}}lar Gu{\ensuremath{\dot{}}}l{\c{c}}ehre, and Bing Xiang. 2016.
\newblock \href {https://doi.org/10.18653/v1/K16-1028} {Abstractive text summarization using sequence-to-sequence {RNN}s and beyond}.
\newblock In \emph{Proceedings of the 20th {SIGNLL} Conference on Computational Natural Language Learning}, pages 280--290, Berlin, Germany. Association for Computational Linguistics.

\bibitem[{Ni et~al.(2025)Ni, Chen, Li, Chen, Li, Wang, Wang, Wang, Zhang, Fan et~al.}]{ni2025surveylargelanguagemodel}
Shiwen Ni, Guhong Chen, Shuaimin Li, Xuanang Chen, Siyi Li, Bingli Wang, Qiyao Wang, Xingjian Wang, Yifan Zhang, Liyang Fan, et~al. 2025.
\newblock \href {https://arxiv.org/abs/2508.15361} {{A Survey on Large Language Model Benchmarks}}.
\newblock \emph{arXiv preprint arXiv:2508.15361}.

\bibitem[{Park et~al.(2024)Park, Lee, Jeong, Park, Koo, Hwang, Park, and Lee}]{parkMultiPragEvalMultilingualPragmatic2024}
Dojun Park, Jiwoo Lee, Hyeyun Jeong, Seohyun Park, Youngeun Koo, Soonha Hwang, Seonwoo Park, and Sungeun Lee. 2024.
\newblock \href {https://snu.elsevierpure.com/en/publications/multiprageval-multilingual-pragmatic-evaluation-of-large-language} {{MultiPragEval: Multilingual Pragmatic Evaluation of Large Language Models}}.
\newblock In \emph{2nd Workshop on Generalisation (Benchmarking) in NLP, GenBench 2024}, pages 96--119. Association for Computational Linguistics (ACL).

\bibitem[{Petroni et~al.(2019)Petroni, Rockt{\"a}schel, Riedel, Lewis, Bakhtin, Wu, and Miller}]{petroni-etal-2019-language}
Fabio Petroni, Tim Rockt{\"a}schel, Sebastian Riedel, Patrick Lewis, Anton Bakhtin, Yuxiang Wu, and Alexander Miller. 2019.
\newblock \href {https://doi.org/10.18653/v1/D19-1250} {{Language Models as Knowledge Bases?}}
\newblock In \emph{Proceedings of the 2019 Conference on Empirical Methods in Natural Language Processing and the 9th International Joint Conference on Natural Language Processing (EMNLP-IJCNLP)}, pages 2463--2473, Hong Kong, China. Association for Computational Linguistics.

\bibitem[{Pilehvar and Camacho-Collados(2019)}]{pilehvar-camacho-collados-2019-wic}
Mohammad~Taher Pilehvar and Jose Camacho-Collados. 2019.
\newblock \href {https://doi.org/10.18653/v1/N19-1128} {{W}i{C}: the word-in-context dataset for evaluating context-sensitive meaning representations}.
\newblock In \emph{Proceedings of the 2019 Conference of the North {A}merican Chapter of the Association for Computational Linguistics: Human Language Technologies, Volume 1 (Long and Short Papers)}, pages 1267--1273, Minneapolis, Minnesota. Association for Computational Linguistics.

\bibitem[{Ponti et~al.(2020)Ponti, Glava{\v{s}}, Majewska, Liu, Vuli{\'c}, and Korhonen}]{ponti-etal-2020-xcopa}
Edoardo~Maria Ponti, Goran Glava{\v{s}}, Olga Majewska, Qianchu Liu, Ivan Vuli{\'c}, and Anna Korhonen. 2020.
\newblock \href {https://doi.org/10.18653/v1/2020.emnlp-main.185} {{XCOPA}: A multilingual dataset for causal commonsense reasoning}.
\newblock In \emph{Proceedings of the 2020 Conference on Empirical Methods in Natural Language Processing (EMNLP)}, pages 2362--2376, Online. Association for Computational Linguistics.

\bibitem[{Raihan et~al.(2025)Raihan, Anastasopoulos, and Zampieri}]{raihan-etal-2025-mhumaneval}
Nishat Raihan, Antonios Anastasopoulos, and Marcos Zampieri. 2025.
\newblock \href {https://doi.org/10.18653/v1/2025.naacl-long.570} {m{H}uman{E}val - a multilingual benchmark to evaluate large language models for code generation}.
\newblock In \emph{Proceedings of the 2025 Conference of the Nations of the Americas Chapter of the Association for Computational Linguistics: Human Language Technologies (Volume 1: Long Papers)}, pages 11432--11461, Albuquerque, New Mexico. Association for Computational Linguistics.

\bibitem[{Rao and Tetreault(2018)}]{rao2018dear}
Sudha Rao and Joel Tetreault. 2018.
\newblock \href {https://aclanthology.org/N18-1012/} {{Dear Sir or Madam, May I Introduce the GYAFC Dataset: Corpus, Benchmarks and Metrics for Formality Style Transfer}}.
\newblock In \emph{Proceedings of the 2018 Conference of the North American Chapter of the Association for Computational Linguistics: Human Language Technologies, Volume 1 (Long Papers)}, pages 129--140.

\bibitem[{Roemmele et~al.(2011)Roemmele, Bejan, and Gordon}]{roemmele2011copa}
Melissa Roemmele, Cosmin~Adrian Bejan, and Andrew~S. Gordon. 2011.
\newblock \href {https://cdn.aaai.org/ocs/2418/2418-10878-1-PB.pdf} {{Choice of Plausible Alternatives: An Evaluation of Commonsense Causal Reasoning}}.
\newblock In \emph{Proceedings of the 2011 AAAI Spring Symposium on Logical Formalizations of Commonsense Reasoning}.

\bibitem[{Rudinger et~al.(2018)Rudinger, Naradowsky, Leonard, and {Van Durme}}]{rudinger-EtAl:2018:N18}
Rachel Rudinger, Jason Naradowsky, Brian Leonard, and Benjamin {Van Durme}. 2018.
\newblock \href {https://aclanthology.org/N18-2002} {Gender bias in coreference resolution}.
\newblock In \emph{Proceedings of the 2018 Conference of the North American Chapter of the Association for Computational Linguistics: Human Language Technologies}, New Orleans, Louisiana. Association for Computational Linguistics.

\bibitem[{Seveso et~al.(2025)Seveso, Potert{\`i}, Federici, Mezzanzanica, and Mercorio}]{seveso-etal-2025-italic}
Andrea Seveso, Daniele Potert{\`i}, Edoardo Federici, Mario Mezzanzanica, and Fabio Mercorio. 2025.
\newblock \href {https://doi.org/10.18653/v1/2025.naacl-long.68} {{ITALIC}: An {I}talian culture-aware natural language benchmark}.
\newblock In \emph{Proceedings of the 2025 Conference of the Nations of the Americas Chapter of the Association for Computational Linguistics: Human Language Technologies (Volume 1: Long Papers)}, pages 1469--1478, Albuquerque, New Mexico. Association for Computational Linguistics.

\bibitem[{Shi et~al.(2023)Shi, Berrebbi, Chen, Chung, Hu, Huang, Chang, Li, Mohamed, yi~Lee, and Watanabe}]{shi2023mlsuperb}
Jiatong Shi, Dan Berrebbi, William Chen, Ho-Lam Chung, En-Pei Hu, Wei-Ping Huang, Xuankai Chang, Shang-Wen Li, Abdelrahman Mohamed, Hung yi~Lee, and Shinji Watanabe. 2023.
\newblock \href {https://doi.org/10.21437/Interspeech.2023-1316} {Ml-superb: Multilingual speech universal performance benchmark}.
\newblock In \emph{Proceedings of the Annual Conference of the International Speech Communication Association, INTERSPEECH 2023}, pages 884--888.

\bibitem[{Singh et~al.(2024)Singh, Romanou, Fourrier, Adelani, Ngui, Vila-Suero, Limkonchotiwat, Marchisio, Leong, Susanto et~al.}]{singhGlobalMMLUUnderstanding2025}
Shivalika Singh, Angelika Romanou, Cl{\'e}mentine Fourrier, David~I Adelani, Jian~Gang Ngui, Daniel Vila-Suero, Peerat Limkonchotiwat, Kelly Marchisio, Wei~Qi Leong, Yosephine Susanto, et~al. 2024.
\newblock \href {http://arxiv.org/abs/2412.03304} {Global {{MMLU}}: {{Understanding}} and {{Addressing Cultural}} and {{Linguistic Biases}} in {{Multilingual Evaluation}}}.
\newblock \emph{arXiv preprint arXiv:2412.03304}.

\bibitem[{Srivastava et~al.(2023)Srivastava, Rastogi, Rao, Shoeb, Abid, Fisch, Brown, Santoro, Gupta, Garriga-Alonso, Kluska, Lewkowycz, Agarwal, Power, Ray, Warstadt et~al.}]{srivastava2023beyond}
Aarohi Srivastava, Abhinav Rastogi, Abhishek Rao, Abu Awal~Md Shoeb, Abubakar Abid, Adam Fisch, Adam~R. Brown, Adam Santoro, Aditya Gupta, Adrià Garriga-Alonso, Agnieszka Kluska, Aitor Lewkowycz, Akshat Agarwal, Alethea Power, Alex Ray, Alex Warstadt, et~al. 2023.
\newblock \href {https://openreview.net/forum?id=uyTL5Bvosj} {Beyond the imitation game: Quantifying and extrapolating the capabilities of language models}.
\newblock \emph{Transactions on Machine Learning Research}.

\bibitem[{Thellmann et~al.(2024)Thellmann, Stadler, Fromm, Buschhoff, Jude, Barth, Leveling, Flores-Herr, K{\"o}hler, J{\"a}kel et~al.}]{thellmann2024multilingualllmevaluationeuropean}
Klaudia Thellmann, Bernhard Stadler, Michael Fromm, Jasper~Schulze Buschhoff, Alex Jude, Fabio Barth, Johannes Leveling, Nicolas Flores-Herr, Joachim K{\"o}hler, Ren{\'e} J{\"a}kel, et~al. 2024.
\newblock \href {https://arxiv.org/abs/2410.08928} {{Towards Multilingual LLM Evaluation for European Languages}}.
\newblock \emph{arXiv preprint arXiv:2410.08928}.

\bibitem[{Vad{\'a}sz and Ligeti-Nagy(2022)}]{vadasz2022winograd}
No{\'e}mi Vad{\'a}sz and No{\'e}mi Ligeti-Nagy. 2022.
\newblock \href {https://akjournals.com/view/journals/2062/69/4/article-p564.xml} {{Winograd schemata and other datasets for anaphora resolution in Hungarian}}.
\newblock \emph{Acta Linguistica Academica}, 69(4):564--580.

\bibitem[{Wang et~al.(2019)Wang, Pruksachatkun, Nangia, Singh, Michael, Hill, Levy, and Bowman}]{wangSuperGLUEStickierBenchmark2020}
Alex Wang, Yada Pruksachatkun, Nikita Nangia, Amanpreet Singh, Julian Michael, Felix Hill, Omer Levy, and Samuel Bowman. 2019.
\newblock \href {https://proceedings.neurips.cc/paper/2019/hash/4496bf24afe7fab6f046bf4923da8de6-Abstract.html} {{{SuperGLUE}}: {{A Stickier Benchmark}} for {{General-Purpose Language Understanding Systems}}}.
\newblock \emph{Advances in neural information processing systems}, 32.

\bibitem[{Wang et~al.(2023)Wang, Chen, Pei, Xie, Kang, Zhang, Xu, Xiong, Dutta, Schaeffer et~al.}]{wang2023decodingtrust}
Boxin Wang, Weixin Chen, Hengzhi Pei, Chulin Xie, Mintong Kang, Chenhui Zhang, Chejian Xu, Zidi Xiong, Ritik Dutta, Rylan Schaeffer, et~al. 2023.
\newblock \href {https://blogs.qub.ac.uk/wp-content/uploads/sites/7/2024/01/A-comprehensive-Assessment-of-Trustworthiness-in-GPT-Models.pdf} {{DecodingTrust: A Comprehensive Assessment of Trustworthiness in GPT Models.}}
\newblock In \emph{NeurIPS}.

\bibitem[{Wang et~al.(2024)Wang, Ma, Zhang, Ni, Chandra, Guo, Ren, Arulraj, He, Jiang et~al.}]{wangMMLUProMoreRobust2024}
Yubo Wang, Xueguang Ma, Ge~Zhang, Yuansheng Ni, Abhranil Chandra, Shiguang Guo, Weiming Ren, Aaran Arulraj, Xuan He, Ziyan Jiang, et~al. 2024.
\newblock \href {https://proceedings.neurips.cc/paper_files/paper/2024/hash/ad236edc564f3e3156e1b2feafb99a24-Abstract-Datasets_and_Benchmarks_Track.html} {{{MMLU-Pro}}: {{A More Robust}} and {{Challenging Multi-Task Language Understanding Benchmark}}}.
\newblock \emph{Advances in Neural Information Processing Systems}, 37:95266--95290.

\bibitem[{Warstadt et~al.(2020)Warstadt, Parrish, Liu, Mohananey, Peng, Wang, and Bowman}]{warstadt-etal-2020-blimp-benchmark}
Alex Warstadt, Alicia Parrish, Haokun Liu, Anhad Mohananey, Wei Peng, Sheng-Fu Wang, and Samuel~R. Bowman. 2020.
\newblock \href {https://doi.org/10.1162/tacl_a_00321} {{BL}i{MP}: The benchmark of linguistic minimal pairs for {E}nglish}.
\newblock \emph{Transactions of the Association for Computational Linguistics}, 8:377--392.

\bibitem[{Warstadt et~al.(2019)Warstadt, Singh, and Bowman}]{warstadt2019neural}
Alex Warstadt, Amanpreet Singh, and Samuel~R Bowman. 2019.
\newblock \href {https://direct.mit.edu/tacl/article-abstract/doi/10.1162/tacl_a_00290/43528} {Neural network acceptability judgments}.
\newblock \emph{Transactions of the Association for Computational Linguistics}, 7:625--641.

\bibitem[{White et~al.(2024)White, Dooley, Roberts, Pal, Feuer, Jain, Shwartz-Ziv, Jain, Saifullah, Dey et~al.}]{whiteLiveBenchChallengingContaminationLimited2025a}
Colin White, Samuel Dooley, Manley Roberts, Arka Pal, Ben Feuer, Siddhartha Jain, Ravid Shwartz-Ziv, Neel Jain, Khalid Saifullah, Sreemanti Dey, et~al. 2024.
\newblock \href {http://arxiv.org/abs/2406.19314} {{{LiveBench}}: {{A Challenging}}, {{Contamination-Limited LLM Benchmark}}}.
\newblock \emph{arXiv preprint arXiv:2406.19314}.

\bibitem[{Wróbel et~al.(2024)Wróbel, {SpeakLeash Team}, and {Cyfronet Team}}]{polish-medical-llm-leaderboard}
Krzysztof Wróbel, {SpeakLeash Team}, and {Cyfronet Team}. 2024.
\newblock Polish medical leaderboard.
\newblock \url{https://huggingface.co/spaces/speakleash/polish_medical_leaderboard}.

\bibitem[{Xuan et~al.(2025)Xuan, Yang, Qi, Zeng, Xiao, Feng, Liu, Xing, Wang, Gao et~al.}]{xuanMMLUProXMultilingualBenchmark2025b}
Weihao Xuan, Rui Yang, Heli Qi, Qingcheng Zeng, Yunze Xiao, Aosong Feng, Dairui Liu, Yun Xing, Junjue Wang, Fan Gao, et~al. 2025.
\newblock \href {http://arxiv.org/abs/2503.10497} {{{MMLU-ProX}}: {{A Multilingual Benchmark}} for {{Advanced Large Language Model Evaluation}}}.
\newblock \emph{arXiv preprint arXiv:2503.10497}.

\bibitem[{Yang et~al.(2021)Yang, Chi, Chuang, Lai, Lakhotia, Lin, Liu, Shi, Chang, Lin, Huang, Tseng, Liu, Huang, Dong, Wang, Ma, Chen, Chang, Lin, Huang, Wu, Hsu, Chen, Cheng, Tsao, Hsieh, and Lee}]{yang2021superb}
Shu-wen Yang, Po-Han Chi, Yung-Sung Chuang, Cheng-I~Jeff Lai, Kushal Lakhotia, Yist~Y. Lin, Andy~T. Liu, Jiatong Shi, Xuankai Chang, Guan-Ting Lin, Wei-Cheng Huang, Wei-Cheng Tseng, Da-Rong Liu, Zili Huang, Shuyan Dong, Shang-Wen~Li Wang, Zhaoheng~Ni Ma, Benjamin Chen, Chih-Liang Chang, Kevin Lin, Wen-Chin Huang, Andy~T. Wu, Po-Chun Hsu, Chun-Lin Chen, Han~Lu Cheng, Yu~Tsao, Hung-Yi Hsieh, and Hung-Yi Lee. 2021.
\newblock \href {https://waseda.elsevierpure.com/ja/publications/superb-speech-processing-universal-performance-benchmark} {Superb: Speech processing universal performance benchmark}.
\newblock In \emph{Proceedings of Interspeech}.

\bibitem[{Zellers et~al.(2019)Zellers, Holtzman, Bisk, Farhadi, and Choi}]{zellersHellaSwagCanMachine2019}
Rowan Zellers, Ari Holtzman, Yonatan Bisk, Ali Farhadi, and Yejin Choi. 2019.
\newblock \href {https://aclanthology.org/P19-1472/} {{HellaSwag: Can a Machine Really Finish Your Sentence?}}
\newblock In \emph{Proceedings of the 57th Annual Meeting of the Association for Computational Linguistics}, pages 4791--4800.

\bibitem[{Zhang et~al.(2018)Zhang, Liu, Liu, Gao, Duh, and Van~Durme}]{zhang2018record}
Sheng Zhang, Xiaodong Liu, Jingjing Liu, Jianfeng Gao, Kevin Duh, and Benjamin Van~Durme. 2018.
\newblock \href {https://arxiv.org/abs/1810.12885} {Record: Bridging the gap between human and machine commonsense reading comprehension}.
\newblock \emph{arXiv preprint arXiv:1810.12885}.

\bibitem[{Zhong et~al.(2024)Zhong, Cheng, Liu, Jiang, Wan, Chu, Murawaki, and Kurohashi}]{zhongEnglishCentricLLMsWhat2024}
Chengzhi Zhong, Fei Cheng, Qianying Liu, Junfeng Jiang, Zhen Wan, Chenhui Chu, Yugo Murawaki, and Sadao Kurohashi. 2024.
\newblock \href {http://arxiv.org/abs/2408.10811} {{Beyond English-centric LLMs: What language do multilingual language models think in?}}
\newblock \emph{arXiv preprint arXiv:2408.10811}.

\end{thebibliography}


\appendix

\section{Appendix}
\label{sec:appendix}

In Tables \ref{tab:llm_benchmarks}, \ref{tab:llm_benchmarks-part2}, \ref{tab:llm_benchmarks_part3} and \ref{tab:llm_benchmarks_part4}, we provide the full proposed taxonomy for LLM benchmarks along with category descriptions and examples.

\begin{table*}[!ht]
\begin{center}
\begin{tabularx}{\textwidth}{|lp{0.3\textwidth}X|}
\hline
Level & Category & Explanation \\
\hline
1 & \textsc{\textbf{Language capabilities}} & Language-related tasks and scenarios that evaluate core linguistic skills of large language models (LLMs), including natural language understanding, generation, and interaction in human language.\\
\hline
1.1 & Natural language understanding and generation & Ability to interpret, process, and produce grammatically correct, coherent and contextually appropriate text.\\
\hdashline
1.1.1 & \quad Comprehension & Understanding text meaning, intent, and details in written or spoken input. E.g., the English question-answering BoolQ \citep{clark-etal-2019-boolq} dataset, ReCoRD (Reading Comprehension with Commonsense Reasoning, \citealp{zhang2018record}) dataset, and Multi-Sentence Reading Comprehension (MultiRC, \citealp{MultiRC2018}) dataset, the English RTE (Recognizing Textual Entailment, \citealp{giampiccolo2007third-rte}) and Hungarian HuRTE \citep{ligeti-nagyHuLUHungarianLanguage2024} datasets, and others. \\
\hdashline
1.1.2 & \quad Context sensitivity & Identifying meaning, tone or style from contextual clues and responding accordingly; maintaining consistency across longer discourse. E.g., the English WiC (Words in Context, \citealp{pilehvar-camacho-collados-2019-wic}) dataset.\\
\hdashline
1.1.3 & \quad Natural language inference & Determining a relationship between sentences, e.g., whether they contradict, entail, or are neutral to each other. E.g., the English CommitmentBank \citep{de2019commitmentbank} dataset and the Hungarian CommitmentBank Corpus (HuCB, \citealp{ligeti-nagyHuLUHungarianLanguage2024}), the Slovene Natural Language Inference Dataset SI-NLI \citep{si-nli-clarin} and the Hungarian anaphora resolution datasets (HuWNLI, \citealp{vadasz2022winograd}).\\
\hdashline
1.1.4 & \quad Summarization & Condensing text while preserving key information and intent. E.g., the English CNN/DailyMail text summarization benchmark \citep{nallapati-etal-2016-abstractive}, and the English BillSum benchmark for legal text summarization \citep{kornilova-eidelman-2019-billsum}.\\
\hline
1.2 & General linguistic competence & Mastering grammar, style, and pragmatic use of language.\\
\hdashline
1.2.1 & \quad Grammar & Producing text that is grammatically correct in terms of syntax, identifying grammatical errors or correcting them. E.g., the Benchmark of Linguistic Minimal Pairs for English (BLiMP, \citealp{warstadt-etal-2020-blimp-benchmark}), the English Corpus of Linguistic Acceptability (CoLA, \citealp{warstadt2019neural}) and Hungarian Corpus of Linguistic Acceptability (HuCOLA, \citealp{ligeti-nagyHuLUHungarianLanguage2024}), and the Polish linguistic and cultural competency benchmark (PLCC, \citealp{dadasEvaluatingPolishLinguistic2025}) dataset.\\
\hdashline
1.2.2 & \quad Style & Identifying the tone, register, or rhetorical style in an input or generating the output in the appropriate tone, register or rhetorical style. E.g., the English Grammarly’s Yahoo Answers Formality Corpus (GYAFC) for benchmarking formality style transfer \citep{rao2018dear}. \\
\hdashline
1.2.3 & \quad Conversation \& pragmatics & Identifying the intent of the written and spoken output and other social aspects of communication, such as politeness, sarcasm, emotions, implicit and figurative meaning. E.g., the multilingual MultiPragEval \citep{parkMultiPragEvalMultilingualPragmatic2024} and the Slovenian SloPragEval pragmatics understanding benchmarks, the English IMDB \citep{maas-etal-2011-learning} and multilingual ParlaSent (\citealp{parlasent-clarin}; \citealp{mochtak2024parlasent}) sentiment identification datasets, English EN-GINCO and Slovenian GINCO automatic genre identification benchmark (\citealp{kuzman2023automatic}; \citealp{ginco-clarin}), and the English Stanford Politeness Corpus \citep{danescu2013computational}.\\
\hline
\end{tabularx}
\caption{Hierarchy of LLM Benchmark Categories (part 1).}
\label{tab:llm_benchmarks}
 \end{center}
\end{table*}

\begin{table*}[!ht]
\begin{center}
\begin{tabularx}{\textwidth}{|lp{0.3\textwidth}X|}
\hline
Level & Category & Explanation \\
\hline
1.3 & Specific linguistic competence & Mastering specific uses of language, such as creative or specialized language.\\
\hdashline
1.3.1 & \quad Creativity & Producing original, novel, or imaginative language (e.g., storytelling, poetry, jokes). \\
\hdashline
1.3.2 & \quad Terminology \& specialized language & Correct use and understanding of domain-specific vocabulary and professional jargon. \\
\hdashline
1.3.3 & \quad Language varieties & Benchmarks evaluating diverse forms of language use, including child language, second-language learners, and atypical communication. \\
\hline
1.4 & Multilinguality & Ability to operate across languages and dialects.\\
\hdashline
1.4.1 & \quad Translation & Rendering meaning from one language to another with fidelity. E.g., the Flores-101 and FLORES-200 datasets \citep{goyal-etal-2022-flores}, the translation tasks in the BenchMax \citep{huangBenchMAXComprehensiveMultilingual2025a} dataset, and the machine translation benchmarks in the Slovenian SloBENCH evaluation framework (\url{https://slobench.cjvt.si}).\\
\hdashline
1.4.2 & \quad Code-switching / multilingual interaction & Ability to detect the use of multiple languages, identify which language is being used, and to mix languages fluidly within discourse.\\ 
\hdashline
1.4.3 & \quad Dialectal capabilities & Understanding or generating dialectal or regional variations. E.g., the DIALECTBENCH~\citep{faisal-etal-2024-dialectbench} dataset.\\
\hline
1.5 & Speech & Understanding and generation of spoken language, mastery of intonation, prosody, and other features unique to speech. E.g., tasks in the English Speech processing Universal PERformance Benchmark (SUPERB, \citealp{yang2021superb}) and its multilingual extension, the Multilingual Speech processing Universal PERformance Benchmark (ML-SUPERB,~\citealp{shi2023mlsuperb}).\\
\hdashline
1.5.1 & \quad Speech understanding & The ability to transcribe speech and extract meaning, intent, and emotion from spoken input. E.g., the Slovenian Speech Recognition benchmark in the SloBENCH evaluation framework (\url{https://slobench.cjvt.si}).\\
\hdashline
1.5.2 & \quad Speech generation & Generating natural-sounding speech with control over prosody, emotion, and speaker features.\\
\hdashline
1.5.3 & \quad Speech-specific interaction & Managing multi-turn spoken dialogue, maintaining conversational grounding, and understanding paralinguistic cues.\\
\hline
1.6 & Cultural competence & Understanding culturally-specific concepts and awareness of cultural context, norms, and values in communication. E.g., the Polish linguistic and cultural competency benchmark (PLCC, \citealp{dadasEvaluatingPolishLinguistic2025}).\\
\hline
1.7 & Multimodality & Mastering tasks across multiple input/output formats (e.g., text, speech, sign language, audiovisual signal). E.g., the Polish Cultural Vision Benchmark (PCVB, \url{https://huggingface.co/spaces/speakleash/Polish_Cultural_Vision_Benchmark}).\\
\hline
\end{tabularx}
\caption{Hierarchy of LLM Benchmark Categories (part 2).}
\label{tab:llm_benchmarks-part2}
 \end{center}
\end{table*}

\begin{table*}[!ht]
\begin{center}
\begin{tabularx}{\textwidth}{|lp{0.3\textwidth}X|}
\hline
Level & Category & Explanation \\
\hline
2 & \textsc{\textbf{Knowledge and reasoning}} & Generating output that is consistent with established facts and real-world knowledge, logical reasoning and problem solving.\\
\hline
2.1 & General knowledge & Broad world knowledge across common domains and established facts. E.g., the English Massive Multitask Language Understanding (MMLU,~\citealp{hendryckstest2021}) and its translations into European languages \citep{thellmann2024multilingualllmevaluationeuropean} and other languages \citep{singhGlobalMMLUUnderstanding2025}, the English Winograd Schema Challenge (WSC, \citealp{levesque2012winograd}), the English WikiFact benchmark \citep{petroni-etal-2019-language}, and others.\\
\hline
2.2 & Domain-specific knowledge & Specialized knowledge in professional or technical areas.\\
\hdashline
2.2.1 & \quad Math & Numerical reasoning, problem solving, and formal proofs. E.g., the English Omni-MATH (Mathematical Reasoning) dataset \citep{gao2024omni}, certain tasks in the English BIG-bench (Beyond the Imitation Game Benchmark, \citealp{srivastava2023beyond}) and LiveBench \citep{whiteLiveBenchChallengingContaminationLimited2025a} benchmarks.\\
\hdashline
2.2.2 & \quad Law & Generating output that is consistent with relevant laws, cases, and regulations, and performing various forms of legal reasoning, including issue-spotting and interpretation. E.g., the English BillSum benchmark \citep{kornilova-eidelman-2019-billsum} for legal text summarization and the English LegalBench benchmark of legal reasoning tasks \citep{guha2023legalbench}.\\
\hdashline
2.2.3 & \quad Finance & Knowledge of economic principles, financial systems, and reasoning over markets, investments, and accounting.\\
\hdashline
2.2.4 & \quad Medicine & Knowledge of biomedical concepts and interpretation of medical information for diagnostics and treatment. E.g., the English MedQA benchmark \citep{jin2021disease} and the Polish medical leaderboard \citep{polish-medical-llm-leaderboard}.\\
\hline
2.3 & Reasoning & Processing and reasoning about information logically, drawing inferences and problem-solving. This includes understanding cause-and-effect scenarios. E.g., the English ReCoRD (Reading Comprehension with Commonsense Reasoning, \citealp{zhang2018record}) dataset, the English Hellaswag \citep{zellersHellaSwagCanMachine2019} dataset and its translation into 20 other European languages EU20-HellaSwag \citep{thellmann2024multilingualllmevaluationeuropean}, the English COPA \citep{roemmele2011copa} dataset and its translations into Hungarian \citep{ligeti-nagyHuLUHungarianLanguage2024}, Croatian (\citealp{ljubevsic2021copa-hr}; \citealp{copa-hr-clarin}), a Slovenian Cerkno dialect \citep{ljubesic-etal-2024-dialect} and other languages and dialects.\\
\hline
2.4 & Agency & Sustained long-term planning, strategic reasoning and consistent decision-making over extended interactions. E.g., certain tasks from the English LiveBench \citep{whiteLiveBenchChallengingContaminationLimited2025a} benchmark.\\
\hline
\end{tabularx}
\caption{Hierarchy of LLM Benchmark Categories (part 3).}
\label{tab:llm_benchmarks_part3}
 \end{center}
\end{table*}

\begin{table*}[!ht]
\begin{center}
\begin{tabularx}{\textwidth}{|lp{0.3\textwidth}X|}
\hline
Level & Category & Explanation \\
\hline
3 & \textsc{\textbf{Safety and trustworthiness}} & Benchmarks that evaluate the model's reliability and its risks, that is, whether it behaves reliably, predictably, ethically, and without harm.\\
\hline
3.1 & Misinformation & Detecting or generating false or misleading information. E.g., the English TruthfulQA dataset \citep{lin2022truthfulqa} and its translation into 20 other European languages EU20-TruthfulQA \citep{thellmann2024multilingualllmevaluationeuropean}.\\ 
\hline
3.2 & Toxicity & Identifying offensive, abusive, or harmful language, and avoiding its use. Includes tasks such as hate speech detection. E.g., the English CivilComments benchmark \citep{borkan2019nuanced}, and the Toxicity benchmark in the English DecodingTrust benchmark collection \citep{wang2023decodingtrust}.\\
\hline
3.3 & Bias & Detection and mitigation of stereotypes or unfair treatment across groups. Includes benchmarks that assess whether a model’s performance on a task is unjustifiably different across different groups and attributes. E.g., the English Winogender Schema Diagnostics dataset \citep{rudinger-EtAl:2018:N18} and the Fairness and Stereotype benchmarks in the English DecodingTrust benchmark collection \citep{wang2023decodingtrust}. \\
\hline
3.4 & Machine ethics & Adherence to ethical principles and moral reasoning in responses, ability to distinguish between moral and immoral actions. E.g., the Machine Ethics benchmark in the English DecodingTrust benchmark collection \citep{wang2023decodingtrust}.\\
\hline
3.5 & Psychological traits & Assessing model's characteristics that are typically associated with human personalities.\\
\hline
3.6 & Robustness & Reliability under adversarial input, noise, or perturbations. E.g., the Adversarial Robustness, Out-of-Distribution Robustness, and Robustness against Adversarial Demonstrations benchmarks in the English DecodingTrust benchmark collection \citep{wang2023decodingtrust}.\\
\hline
3.7 & Data governance & Benchmarks that assess whether models leak their training data in their responses. E.g., the Privacy benchmark in the English DecodingTrust benchmark collection \citep{wang2023decodingtrust}.\\
\hline
4 & \textsc{\textbf{Alignment and dangerous capabilities}} & Benchmarks that evaluate model capabilities that could have catastrophic consequences, including dangerous capabilities (cyber operations, weapons acquisition, autonomous replication) and alignment risks (power-seeking behavior, shutdown resistance). This category includes tasks that assess whether the model is able to detect and exploit vulnerabilities in hardware, software and data, whether the model can contribute to the development of weapon technologies, whether the model possesses information about itself and its creators, whether it can break free from its operational confines, or whether it is capable of persuasion and manipulation. \\
\hline
\end{tabularx}
\caption{Hierarchy of LLM Benchmark Categories (part 4).}
\label{tab:llm_benchmarks_part4}
 \end{center}
\end{table*}

\end{document}